# DanceCrafter: Fine-Grained Text-Driven Controllable Dance Generation via Choreographic Syntax

Hang Yuan[1,4,5*], Xiaolin Hu[3,5*], Yan Wan[2], Menglin Gao[2], Wenzhe Yu[2], Cong Huang[6], Fei Xu[2], Qing Li[2], Christina Dan Wang[4†], Zhou Yu[1†], Kai Chen[6†]

[1]East China Normal University    [2]Beijing Dance Academy    [3]Beijing University of Posts and Telecommunications
[4]New York University Shanghai    [5]Zhongguancun Academy    [6]Zhongguancun Institute of Artificial Intelligence
*Joint First Authors    †Joint Corresponding Authors

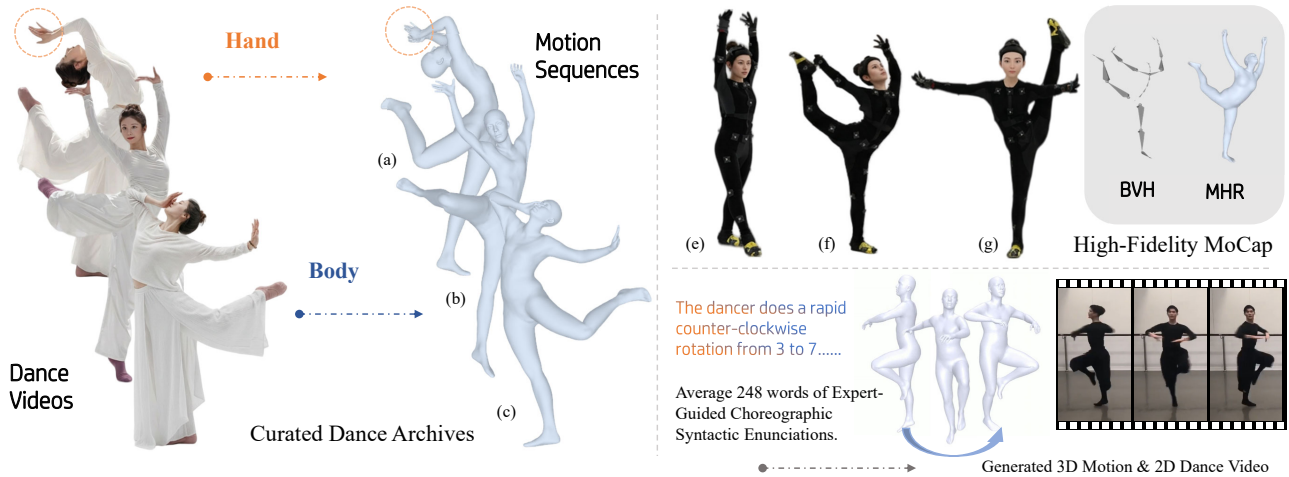

Figure 1: DanceCrafter enables fine-grained text-driven generation of 3D dance motions and expressive 2D videos. We construct DanceFlow, the finest-grained dance dataset to date, grounded in a novel *Choreographic Syntax*. Our data originates from two sources: (Left) curated in-the-wild and professional video archives and (Upper Right) high-fidelity motion capture from dance experts, (Lower Right) Driven by expert-guided, highly detailed choreographic descriptions (averaging 248 words), our tailored generation framework achieves precise control and high-fidelity synthesis of complex dance sequences.

## Abstract

Text-driven controllable dance generation remains under-explored, primarily due to the severe scarcity of high-quality datasets and the inherent difficulty of articulating complex choreographies. Characterizing dance is particularly challenging owing to its intricate spatial dynamics, strong directionality, and the highly decoupled movements of distinct body parts. To overcome these bottlenecks, we bridge principles from dance studies, human anatomy, and biomechanics to propose *Choreographic Syntax*, a novel theoretical framework with a tailored annotation system. Grounded in this syntax, we combine professional dance archives with high-fidelity motion capture data to construct **DanceFlow**, the most fine-grained dance dataset to date. It encompasses 41 hours of high-quality motions paired with 6.34 million words of detailed descriptions. At the model level, we introduce **DanceCrafter**, a tailored motion transformer built upon the Momentum Human Rig. To circumvent optimization instabilities, we construct a continuous manifold motion representation paired with a hybrid normalization strategy. Furthermore, we design an anatomy-aware loss to explicitly regulate the decoupled nature of body parts. Together, these adaptations empower DanceCrafter to achieve the high-fidelity and stable generation of complex dance sequences. Extensive evaluations and user studies demonstrate our state-of-the-art performance in motion quality, fine-grained controllability, and generation naturalness. Project page: https://faustrazor.github.io/.

## 1 Introduction

While human motion data is central to diverse digital applications [26, 31, 43], acquiring high-quality 3D motion via traditional studio capture remains prohibitively expensive. In recent years, modern generative approaches [11, 35, 41, 45] have rapidly advanced, demonstrating powerful natural language controllability that enables users to synthesize fine-grained, diverse movements through intuitive textual descriptions [16, 30]. This generative paradigm is particularly vital for dance. As both an artistic expression and cultural heritage, dance conveys deep emotions through body language, yet producing customized dance content entails prohibitive financial and time costs. The conventional workflow demands expert choreographers to design movements, professional dancers to perform them, and extensive motion capture post-processing. Consequently, 3D dance generation has emerged as a highly promising alternative to empower dance creation.

Most existing works rely on music as the primary control condition to synthesize rhythm-synchronized movements [5, 31, 32]. However, this inherently stochastic paradigm falls short for professional choreography applications, which demand fine-grained, deterministic control over specific movement sequences. As a promising way, text-driven dance generation is severely bottlenecked by data scarcity. Most existing textual motion datasets predominantly feature general everyday actions paired with overly simplistic descriptions. Fine-grained textual dance datasets remain profoundly lacking, primarily because the extreme spatiotemporal complexity and high kinematic degrees of freedom inherent in dance make these movements exceptionally difficult to characterize through natural language. Additionally, most motion generation methods are built upon the SMPL and SMPL-X parametric body models [25, 27], which couple skeletal posture with surface geometry and provide strong human-body priors. However, this entanglement poses challenges in specialized domains like dance, which features highly decoupled, large-amplitude limb movements. Consequently, complex dance executions may trigger structural artifacts, such as the "candy-wrapper effect" [8, 19]. In contrast, the recent Momentum Human Rig (MHR) [8] introduces a decoupled modeling paradigm, effectively mitigating these issues.

To overcome these critical bottlenecks, we propose the **DanceFlow** dataset. We establish this robust data foundation by first collecting a large corpus of dance videos from the internet and the archives of a professional dance academy. Furthermore, we capture a set of professional dancers' performances using an optical motion capture system [24], thereby ensuring motion accuracy and providing high-precision ground truth. The collected data is subsequently processed through a rigorous data pipeline. In total, we assemble **41** hours of dance data comprising over **20K** motion segments. We then systematically analyze the core challenges inherent in textual dance description. By integrating interdisciplinary frameworks across choreography, anatomy, and biomechanics, we formulate *Choreographic Syntax*, a theoretical framework for dance motion description comprising four core dimensions: Body, Space, Orientation, and Effort. Leveraging this syntax, we construct **DanceFlow**, the most fine-grained text-annotated dance dataset to date. The dataset comprises a total of **6.34M** words of detailed textual descriptions. With an average of **248** words per motion sequence, it substantially surpasses the current SOTA (48 words) [16]. Built upon this extensive dataset, we propose **DanceCrafter**, a text-driven dance generation framework featuring a tailored motion transformer based on the MHR. To effectively mitigate optimization instabilities, we formulate a continuous manifold representation for the motion data, complemented by a hybrid normalization mechanism. Furthermore, to explicitly govern the highly decoupled movements of varying body parts during complex dance routines, we introduce a novel anatomy-aware objective function. Synergistically, these targeted algorithmic enhancements enable DanceCrafter to precisely align fine-grained textual instructions with abstract 3D dance concepts, ultimately yielding high-fidelity and stable motion sequences. However, pure 3D motions lack the visual richness and costume details of real dance performances. To bridge this gap, we cascade a video generation model to synthesize expressive dance videos.

In summary, our main contributions are as follows:

- **Choreographic Syntax & Large-Scale Dataset**: We collect 41 hours of professional dance data and design a Choreographic Syntax grounded in dance theory, anatomy, and biomechanics. Leveraging this syntax, we construct **DanceFlow**, the most fine-grained text-annotated dance dataset to date. It contains over 6 million words words of detailed descriptions, substantially surpassing the previous SOTA.
- **DanceCrafter Framework**: We propose **DanceCrafter**, a text-driven dance generation framework that specifically leverages continuous manifold representations and hybrid regularizations within the MHR. Furthermore, through a cascaded system, we achieve the generation of high-quality and controllable 3D dance motion and video.
- **Comprehensive Evaluation**: Extensive quantitative and qualitative experiments demonstrate our method's effectiveness. Ablation studies further validate the contribution of the Choreographic Syntax and the scalability of fine-grained text descriptions for controllable dance generation.

## 2 Related Work

### 2.1 3D Dance Dataset and Parametric Model

Existing dance datasets [18–20] primarily target music-to-dance generation. In contrast, text-annotated datasets mostly cover general motions with coarse descriptions (e.g., 9–12 words in HumanML3D [12] and Motion-X [21]). Even the expert-annotated duet dance dataset, MDD [13], averages only 41 words. Since richer textual granularity significantly boosts generation quality [16], developing fine-grained text annotations for dance is critically needed.

Regarding 3D human parametric models, dominant SMPL [25] and SMPL-X [27] couple posture and shape within a single parameter space. This entanglement often causes artifacts like foot skating and body-hand desynchronization in complex dances [8, 19]. The MHR [8] resolves this by decoupling posture and shape into independently controllable spaces, providing coherent continuous representations ideal for dance. Concurrently, SOTA 3D human reconstruction methods like SAM3D-body [42] enable high-quality 3D motion extraction, facilitating our large-scale dataset construction.

### 2.2 Motion Generation

Text-driven human motion generation has progressed rapidly across diverse generative paradigms. Early approaches adopted GANs and VAEs for motion synthesis [12], while diffusion-based methods such as MDM [35] and MotionDiffuse [45] brought substantial quality improvements by modeling the denoising process directly in motion space. Subsequent works explored latent-space diffusion (MLD [4]), autoregressive token prediction (T2M-GPT [44]), and masked generative modeling (MoMask [11]). More recently, flow matching [22, 23] has emerged as a compelling alternative that learns continuous-time velocity fields via simple regression objectives, offering straighter sampling trajectories and faster inference. These methods are predominantly text-conditioned and trained on general everyday motion datasets. HY-Motion [41], currently the largest text-to-motion model with 1B parameters, demonstrates that scaling both data volume and model size can yield superior generation quality. SnapMoGen [16] further shows that scaling the granularity of textual descriptions, rather than just data volume, significantly improves instruction-following capability and controllability. However, these general-purpose methods are primarily designed for simple everyday actions and perform poorly on highly dynamic dance tasks, where movements exhibit extreme spatiotemporal complexity and high kinematic degrees of freedom.

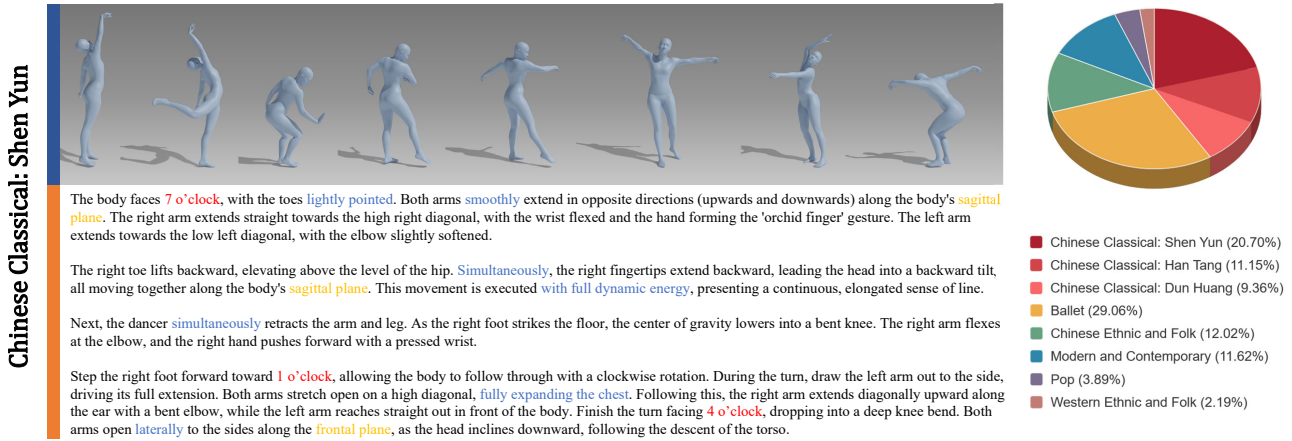

Figure 2: Overview of the dance category composition in our dataset. The figure summarizes the major dance categories and their proportions, and further presents a representative dance example together with fine-grained choreographic description.

Dance generation has been explored predominantly under music conditioning. Recent methods [5, 19, 20, 32, 37] produce increasingly realistic dance sequences driven by musical beats and rhythmic features. While these approaches excel at generating rhythmically synchronized movements, they offer limited semantic controllability, and users cannot specify *what* movements to perform, only the musical context. This makes music-conditioned generation inherently stochastic and unsuitable for professional choreography applications such as film production or stage performances, where dancers must execute precisely specified movements according to a director's vision. TM2D [9] and MDD [13] attempt to incorporate text as an additional modality, but their textual annotations remain coarse (averaging 41 words), far from sufficient to capture the action-by-action intricacies of professional dance. Our work addresses this gap by introducing a Choreographic Syntax that enables unprecedented textual granularity for dance description, and by designing a DiT-based flow matching architecture specifically tailored for the MHR parameter space with anatomy-aware supervision and manifold-preserving normalization.

### 2.3 Dance Theory, Anatomy, and Biomechanics

Theoretical frameworks from dance disciplines and human anatomy play a critical role in formalizing movement description. Laban's concept of "Effort" [39] characterizes the dynamic quality and inner intention of human movement across four dimensions: weight, space, time, and flow. To address the fundamental problem of spatial orientation, Laban also introduced the theory of Choreutics (Space Harmony) [40], which establishes a geometric framework for positioning human motion in space. Complementing these spatial concepts, Vaganova [38], a pioneering classical ballet educator, formulated a definitive system of eight spatial directions specifically tailored for stage dancers. From a biomechanical perspective, Calais Germain [2] proposed a comprehensive anatomical segmentation system of the human body. Together, these multidisciplinary theories—spanning spatial positioning, motion dynamics, and anatomical structure—provide the theoretical foundation for the fine-grained Choreographic Syntax developed in our work.

## 3 The DanceFlow Dataset

Comprising 36 hours of curated video reconstructions and 5 hours of high-precision motion capture, **DanceFlow** offers over 20K processed motion segments. These are paired with 6.34 million fine-grained choreographic words, establishing the most detailed text-motion dataset to date. Figure 2 illustrates the diverse distribution of dance categories within our dataset, alongside a sample.

### 3.1 Data Collection and Processing

We initially collect ~100 hours of dance videos from professional academy archives and the internet. To ensure data quality, we employ a rigorous filtering pipeline: first, we discard low-resolution footage and use PySceneDetect [3] to eliminate discontinuities or viewpoint transitions. Next, a VLM-based filter (Qwen-3.5-Plus [34]) excludes instructional tutorials, multiple subjects, partial-body close-ups, and subtitle occlusions (see Appendix). This yields 36 hours of high-quality, single-person dance videos, from which we extract 3D MHR parameters [8] using SAM3D-body [42]. Furthermore, to mitigate video-estimation inaccuracies, we augment the dataset with 5 hours of optical motion capture from professional dancers, properly retargeted to the MHR space.

Table 1: Comparison of Text-to-Motion Generation Datasets. Note that MDD and our DanceFlow specifically focus on dance.

| Dataset | Domain | Annotation Paradigm | Duration | MoCap | 3D Rep. | # Texts | Avg. Words |
|---|---|---|---|---|---|---|---|
| KIT-ML[29] | General Motion | General Annotators | 10.3h | ✓ | - | 6,278 | 8 |
| HumanML3D [12] | General Motion | General Annotators | 28.6h | ✓ | SMPL | 44,970 | 12 |
| Motion-X[21] | General Motion | Gen. Annotators + Video-LLaMA | 144.2h | ✗ | SMPL-X | 81,084 | 9 |
| SnapMoGen [16] | General Motion | Gen. Annotators + ChatGPT | 43.7h | ✓ | - | 122,565 | 48 |
| MDD [13] | Duet Dance | Domain Experts + GPT-4o | 10.34h | ✓ | SMPL-X | 10,187 | 41 |
| **DanceFlow (Ours)** | **Prof. Dance** | **Domain Experts + Gemini-3-pro-preview** | **41.0h** | ✓ | **MHR** | **6,347,849** | **248** |

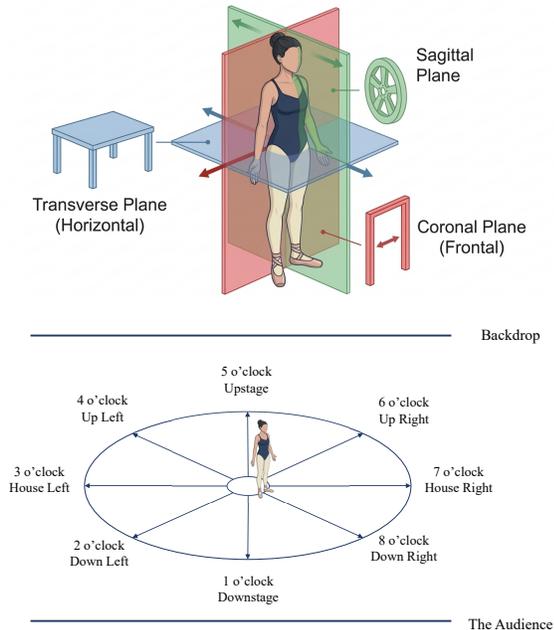

Figure 3: The Space and Orientation dimensions of our Choreographic Syntax. (Top) Movement trajectories are mapped across three anatomical planes. (Bottom) Eight spatial directions are used to anchor the dancer's body orientation.

## 3.2 Choreographic Syntax

Precisely describing dance movements in natural language poses three fundamental challenges. **(1) Ambiguous spatiotemporal dynamics**: Dance entails continuous spatial trajectories and intricate temporal rhythms that simple phrasing struggles to capture. Terms like "move the arm up" fail to convey the exact path, speed, and acceleration profile of a gesture. **(2) Strong directionality in 3D space**: Dance movements are inherently three-dimensional. Naive descriptions based on a single viewpoint (e.g., "left" or "right") are ambiguous, hindering precise spatial targeting. **(3) High decoupling and asymmetry of limb movements**: During dance, different body segments operate with extreme independence; within a motion, the upper and lower limbs may execute entirely distinct movements simultaneously. Conventional language easily causes information loss and confusion when attempting to describe such parallel, asymmetric coordination.

To overcome these challenges, we draw upon interdisciplinary theoretical frameworks spanning choreographic theory, anatomy, biomechanics and kinesiology. Inspired by the core principles of structural linguistics[40], we pioneer a systematic *Choreographic Syntax*. This system scientifically deconstructs complex dance movements into four core dimensions: **Body** (anatomical segments), **Space** (spatial trajectories), **Orientation** (body directions), and **Effort** (dynamic qualities), with fine-grained modeling for each dimension. Specifically, we partition the human body into independent anatomical modules, including the head, upper limbs, trunk (back, waist, abdomen), and lower limbs, enabling the parallel description of each segment's motion. As illustrated in Figure 3, spatial paths are mapped onto three anatomical planes (Transverse, Sagittal, and Frontal) alongside the six fundamental directions of the body's kinesphere. This decomposition resolves common linguistic ambiguities. For instance, a generic "leg kick" is highly ambiguous; as shown on the right of Figure 1, pose (f) depicts a backward kick along the wheel plane, while pose (g) depicts a rightward kick along the Coronal plane. Conversely, pose (b) illustrates a bent-knee upward lift occupying the transverse plane at hip level. We precisely anchor the dancer's orientation using eight spatial directions [38]. For effort dynamics, we adopt Laban's four dimensions (Weight, Space, Time, and Flow) to capture qualitative textures like "sustained" or "explosive." Through this hierarchical deconstruction, we transform continuous dance into a structured description system with extremely high semantic density. Detailed formulations of *Choreographic Syntax* are provided in the Appendix.

## 3.3 Annotation Pipeline

We develop a scalable annotation paradigm by synergizing our Choreographic Syntax with Gemini-3-pro-preview [10] (hereafter Gemini). We first formulate the syntax into structured prompts, directing the model to annotate each motion segment across the four established dimensions. To ensure robust in-context learning, these prompts are augmented with detailed guidelines and expert-curated reference annotations. To guarantee annotation fidelity at scale, we adopt a statistical quality control framework [17]. The 20K segments are divided into 100 batches, with $n$=30 samples per batch randomly drawn for review. Experts rate each on a 5-point scale; scores below 3 are deemed unacceptable. Batches falling below a 95% acceptance rate are re-annotated until compliant. This rigorous pipeline ensures the reliable translation of raw movement into structured, semantically dense choreographic descriptions.

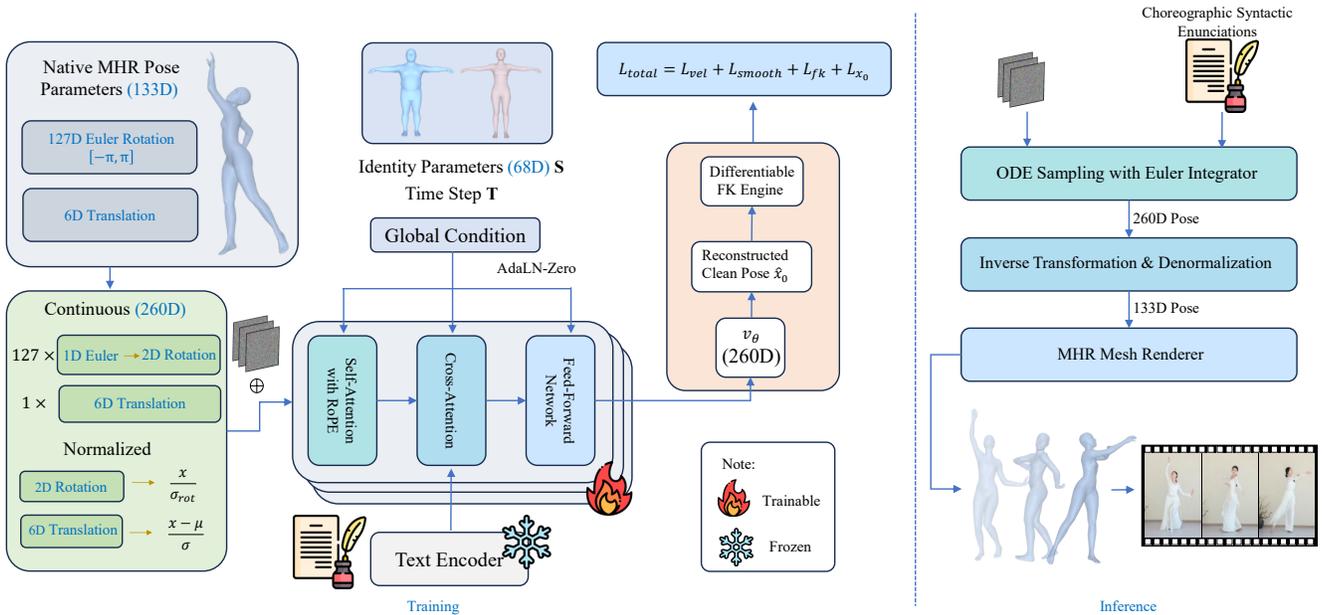

Figure 4: Overview of the DanceCrafter framework. (Left) Training Flow: Native MHR parameters are converted to a continuous representation and processed via hybrid normalization. A DiT backbone learns a conditional velocity field supervised by tailored losses, where choreographic text is injected via cross-attention, and identity and timestep are modulated via adaLN-Zero. (Right) Inference Flow: The generated motion is inverse-transformed to the native MHR space and cascaded with a video expert to synthesize high-fidelity 3D and expressive dance videos.

## 4 Method

To achieve high-fidelity and controllable dance generation, DanceCrafter systematically integrates continuous motion representations and tailored objectives. As illustrated in Figure 4, we first detail our continuous manifold data representation (Section 4.1), followed by the conditional flow matching architecture (Section 4.2), and conclude with our specific training losses and inference procedures (Section 4.3).

### 4.1 Motion Representation and Normalization

We utilize the MHR [8] as our 3D human parametric model. A single MHR frame is characterized by a 204-dimensional vector: a 68-dimensional identity parameter $\mathbf{s} \in \mathbb{R}^{68}$ controlling static body shape, and a 136-dimensional pose parameter capturing dynamic motion. Over a sequence length $T$, the pose parameter $\mathbf{x}_{mhr} \in \mathbb{R}^{T \times 136}$ comprises 6 root translation parameters, alongside 3 global and 127 local Euler rotation angles (with 3 jaw parameters zeroed by default). While directly regressing these Euler angles is intuitive, we observe it induces severe gimbal lock and temporal jittering artifacts. This instability fundamentally stems from the topological incompatibility of a network mapping its Euclidean output space directly to the $SO(3)$ rotation group, causing discontinuous jumps when angles wrap around boundaries (e.g., from $\pi$ to $-\pi$). To formulate a strictly continuous mapping, we convert the orientation of multi-DoF joints into a 6D rotation representation [46]. Specifically, we map the Euler angles to a $3 \times 3$ rotation matrix and extract its first two column vectors, generating a smooth $\mathbb{R}^6$ manifold that gracefully circumvents gimbal lock. Synchronously, we encode each 1-DoF angle $\theta$ as a continuous sine-cosine pair $(\cos \theta, \sin \theta)$. This joint transformation definitively eliminates discontinuous boundary wrap-arounds, bounds all rotational values within $[-1, 1]$, and expands our strictly continuous pose representation to $\mathbf{x}_0 \in \mathbb{R}^{T \times 260}$, significantly stabilizing the generative training process. Because these converted rotation-like features strictly reside on geometric manifolds (i.e., spherical for 6D rotations and circular for sine-cosine pairs), applying standard normalization across all dimensions would destroy their intrinsic topology. Therefore, we introduce a *hybrid normalization* strategy. We divide the rotation dimensions by a single global standard deviation $\sigma_{rot}$ to preserve the manifold structures, while exclusively subjecting the 6 root translation dimensions to standard per-dimension mean-variance normalization. The resulting normalized representation $\bar{\mathbf{x}}_0$ serves as our stable geometric target in flow matching.

### 4.2 Conditional Flow Matching

We adopt the flow matching framework [22, 23] to learn a conditional velocity field transporting Gaussian noise $\mathbf{x}_1 \sim \mathcal{N}(\mathbf{0}, \mathbf{I})$ to the normalized motion distribution $\bar{\mathbf{x}}_0 \in \mathbb{R}^{T \times 260}$. Using a linear interpolation path $\mathbf{x}_t = (1-t)\bar{\mathbf{x}}_0 + t\mathbf{x}_1$ for $t \in [0, 1]$, the velocity network $\mathbf{v}_\theta$ is trained to predict the vector field $\mathbf{x}_1 - \bar{\mathbf{x}}_0$ via the standard objective:

$$\mathcal{L}_{fm} = \mathbb{E}_{t \sim \mathcal{U}(0,1), \bar{\mathbf{x}}_0, \mathbf{x}_1} \left\| \mathbf{v}_\theta(\mathbf{x}_t, t, \mathbf{s}, \mathbf{y}) - (\mathbf{x}_1 - \bar{\mathbf{x}}_0) \right\|^2. \quad (1)$$

Our generative backbone is a Diffusion Transformer (DiT) [28] built upon stacking self-attention, cross-attention, and feed-forward layers. To robustly encode temporal ordering, we apply Rotary Position Embedding (RoPE) [33] coupled with QK-Norm [14] during self-attention, ensuring the stable integration of relative positional geometry.

The choreographic text is encoded by a frozen UMT5 encoder [7] into token-level embeddings $\mathbf{y}$. Through cross-attention, each frame dynamically attends to relevant text segments, which is a structural necessity given our fine-grained dense annotations. Concurrently, the timestep $t$ and body identity parameter $\mathbf{s}$ are projected and summed to form a global conditioning vector, which modulates every Transformer block via AdaLN-Zero [28]. Following standard practices, we employ Classifier-Free Guidance (CFG) [15] during inference to enhance text-motion alignment, enabled by randomly dropping the text and identity conditions during training.

### 4.3 Training Losses and Inference

**Anatomy-Aware Velocity Loss.** To explicitly govern the highly decoupled nature of professional dance movements, we go beyond the base flow matching objective $\mathcal{L}_{\text{fm}}$ by decomposing the 260-dimensional velocity prediction into three distinct anatomical groups: global rotation, structural body joints, and hand joints. We apply group-specific MSE weighting:

$$\mathcal{L}_{\text{vel}} = \lambda_{\text{rot}} \mathcal{L}_{\text{rot}} + \lambda_{\text{body}} \mathcal{L}_{\text{body}} + \lambda_{\text{hand}} \mathcal{L}_{\text{hand}}. \quad (2)$$

Notably, we assign the highest weight to the body subset to strictly prioritize the large-amplitude limb and torso movements central to dance choreography.

**Auxiliary Losses.** We formulate the reconstructed clean pose as $\hat{\mathbf{x}}_0 = \mathbf{x}_t - t \cdot \mathbf{v}_\theta$. To enforce high-fidelity temporal coherence, we apply a direct reconstruction objective ($\mathcal{L}_{x_0} = \lambda_{x_0} \|\hat{\mathbf{x}}_0 - \mathbf{x}_0\|^2$) alongside velocity and acceleration smoothing terms:

$$\mathcal{L}_{\text{smooth}} = \lambda_v \|\Delta \hat{\mathbf{x}}_0 - \Delta \bar{\mathbf{x}}_0\|^2 + \lambda_a \|\Delta^2 \hat{\mathbf{x}}_0 - \Delta^2 \bar{\mathbf{x}}_0\|^2, \quad (3)$$

where $\Delta$ and $\Delta^2$ compute the first and second-order temporal finite differences. Crucially, to constrain physical realism, we inverse-transform $\hat{\mathbf{x}}_0$ back into the 136D MHR space and apply a differentiable forward kinematics (FK) module. This yields robust 3D joint positions, enabling us to enforce a comprehensive kinematic loss $\mathcal{L}_{\text{fk}}$ covering precise joint positioning, linear velocity, and rigid foot-ground contact (detailed in Appendix). Our final training objective explicitly combines these targets:

$$\mathcal{L}_{\text{total}} = \mathcal{L}_{\text{vel}} + \mathcal{L}_{x_0} + \mathcal{L}_{\text{smooth}} + \mathcal{L}_{\text{fk}}, \quad (4)$$

where individual weighting coefficients are absorbed into their respective definitions.

**Inference and Cascaded Video Generation.** During inference, we initialize Gaussian noise $\mathbf{x}_1 \sim \mathcal{N}(\mathbf{0}, \mathbf{I})$ and iteratively solve the probability flow ODE backward to $t = 0$ via a standard Euler integrator mapping $\mathbf{x} \leftarrow \mathbf{x} + \Delta t \cdot \mathbf{v}_{\text{guided}}$, where $\mathbf{v}_{\text{guided}}$ denotes the classifier-free guidance adjusted velocity field. The finalized sample is denormalized and mathematically inverted back to the native 136D MHR pose. Finally, the 3D meshes rendered from the pose parameters provide motion conditioning for Wan-Animate [6] to animate a single reference character image into photorealistic dance videos with the same motion.

## 5 Experiments

### 5.1 Experimental Setup

*Datasets.* We construct our test set via stratified sampling from both our motion capture (MoCap) and video-reconstructed datasets. By proportionally sampling sequences from each source, we ensure a balanced representation of high-precision MoCap sequences and diverse, in-the-wild motions, yielding a total of 1,100 test sequences. Given that existing benchmarks for text-driven dance generation remain scarce, and prevailing text-to-motion datasets (e.g., HumanML3D [12]) primarily feature general human behaviors rather than specialized professional dance, we conduct our evaluations exclusively on this newly curated test set to accurately reflect the unique characteristics of dance movements.

*Evaluation Metrics.* Following standard practices in text-to-motion generation, we adopt two mainstream evaluation protocols: HumanML3D [12] and AIST++ [18]. Their feature extraction methods differ significantly, as HumanML3D uses a learned motion-text encoder while AIST++ relies on rule-based kinetic and geometric features. Consequently, we report results under both protocols for a comprehensive and fair assessment. These protocols evaluate model performance across three key dimensions: (1) **Generation Quality**: For HumanML3D, we report the Fréchet Inception Distance (**FID**) to measure the distributional discrepancy between generated and ground-truth embeddings. For AIST++, we report $\text{FID}_k$ and $\text{FID}_g$, corresponding to its kinetic and geometric feature spaces. (2) **Diversity**: We report **Diversity** (HumanML3D) and $\text{Dist}_k$, $\text{Dist}_g$ (AIST++), computed as the average pairwise Euclidean distance within sets of generated motions to capture motion variation. (3) **Instruction Following**: Under the HumanML3D protocol, we report MultiModal Distance (**MM Dist**), which evaluates text-motion semantic alignment via the Euclidean distance between their respective embeddings.

*Baselines.* As specialized text-to-dance generation remains largely underexplored, we compare our approach against general text-to-motion generation models, including T2M [12], MDM [36], MoMask [11], and HY-Motion [41]. Additionally, we compare our method against TM2D [9], a dance generation method conditioned on both music and text. To ensure a fair comparison within our text-only paradigm, we supply TM2D with an empty music condition during inference. Furthermore, since both the baseline methods and the evaluation protocols natively operate on SMPL-X representations, we uniformly convert our model's generated MHR motion parameters into the SMPL-X format prior to metric computation.

*User Study.* To complement our quantitative results, we conduct a user study comparing our method against the aforementioned baselines. We randomly sample 3 diverse text prompts from the test set and generate corresponding dance sequences using all evaluated models. The generated motions are subsequently rendered into videos for a side-by-side perceptual evaluation. We recruit 20 participants who rate each video on a 5-point Likert scale (1=worst, 5=best) assessing three criteria: (1) **Text-Motion Alignment**, measuring how well the generated dance semantically aligns with the input prompt; (2) **Motion Quality**, evaluating the overall quality of the generated dance motion; and (3) **Aesthetic Appeal**, assessing how aesthetically pleasing and expressive the generated dance is.

Table 2: Quantitative comparison against baseline methods across two widely adopted protocols. For readability, the reported AIST++ FID values are scaled by $10^{-2}$. Arrows indicate whether a lower (↓), or closer-to-real (→) value is better. Best and second-best results are bolded and underlined, respectively.

| Method | HumanML3D Protocol | | | AIST++ Protocol | | | |
|---|---|---|---|---|---|---|---|
| | FID ↓ | MM Dist → | Diversity → | $FID_k$ ↓ | $FID_g$ ↓ | $Dist_k$ → | $Dist_g$ → |
| Real Data | - | 4.630 | 2.886 | - | - | 11.086 | 7.050 |
| T2M [CVPR'22] | 35.188 | 3.910 | <u>2.315</u> | <u>1.879</u> | 1.465 | **9.480** | 26.048 |
| MDM [ICLR'23] | 11.519 | 5.791 | 5.551 | 9.276 | 13.555 | 36.608 | 34.956 |
| TM2D [ICCV'23] | 14.509 | 6.628 | 5.593 | 3.394 | 32.691 | 5.602 | 12.537 |
| MoMask [CVPR'24] | <u>7.424</u> | 5.762 | 4.947 | 7.244 | <u>7.843</u> | 31.592 | 27.740 |
| HY-Motion [Arxiv'25] | 17.826 | <u>4.918</u> | 5.510 | 18.398 | 29.410 | 33.482 | 50.217 |
| **Ours** | **0.868** | **4.476** | **2.909** | **0.273** | **0.150** | <u>7.334</u> | **5.088** |

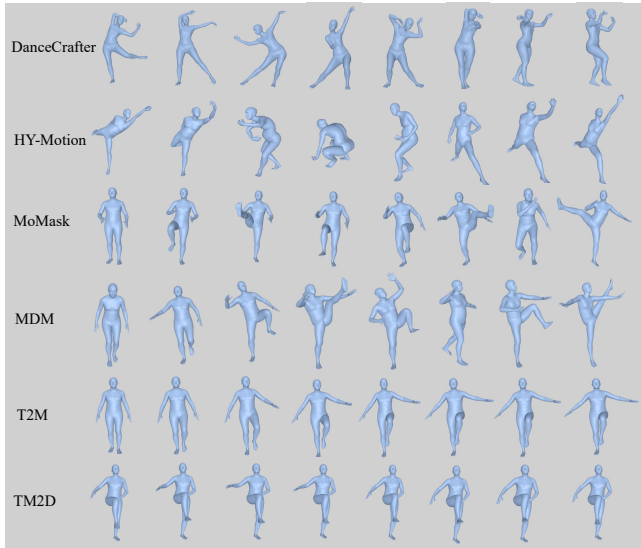

Figure 5: Qualitative comparison against baseline methods. We visualize uniformly sampled frames from sequences generated using the same text prompt. Our method produces more coherent and expressive dance poses.

### 5.2 Main Results

Table 2 reports the quantitative comparison under two evaluation protocols. Figure 6 shows the average user-study scores based on 3 randomly selected test samples, while Figure 5 presents a qualitative visualization of 1 sample selected from these 3 samples.

Under the HumanML3D protocol, our method achieves the best overall performance, yielding the lowest FID (0.868) and substantially outperforming all baselines. It also secures the best MM Dist (4.476) and a Diversity score (2.909) nearly identical to the ground-truth value (2.886). Conversely, most baselines struggle with our detailed dance descriptions, likely due to the scarcity of fine-grained dance-text pairs in existing training datasets. Consequently, they often exhibit inflated diversity metrics, indicating a distributional drift from real dance motions. Among the baselines, MoMask and MDM achieve the most competitive FID scores (7.424 and 11.519, respectively). As reflected in Figure 5, both models appear relatively stable: while they struggle to faithfully synthesize unseen fine-grained textual details, they successfully capture several salient motion cues. However, their high MM Dist scores (5.762 and 5.791) highlight limited text-motion alignment under long-form descriptions. In contrast, T2M performs much worse (FID: 35.188), frequently manifesting jittering and collapsing into repetitive patterns, as further evidenced by an abnormally low Diversity score (2.315). TM2D generates comparatively monotonous or near-static motions, revealing that without its native music condition, the text modality alone is insufficient for effective control. Finally, while HY-Motion achieves a favorable MM Dist (4.918), its FID remains high (17.826); qualitative observations reveal abnormal body twisting and inter-penetration, a likely artifact of the model misinterpreting complex, unseen textual instructions.

Under the AIST++ protocol, our method consistently demonstrates superior performance. By achieving the lowest $FID_k$ (0.273) and $FID_g$ (0.150), it exhibits the strongest distributional alignment with real motions in the dance-specific feature space. Furthermore, our model maintains Diversity metrics ($Dist_g$: 5.088, $Dist_k$: 7.334) that most closely approximate real-data statistics. For the baselines, performance trends remain largely consistent with HumanML3D. MoMask and MDM remain relatively stable, ranking among the better general-purpose models. TM2D exhibits a particularly high $FID_g$ (32.691), reaffirming the severe degradation caused by the absent music modality. HY-Motion's poor metrics corroborate our earlier visual observations of unphysical body distortions. Interestingly, although T2M shows numerical improvement under AIST++, our qualitative analysis confirms that its generated motions still suffer from instability and jittering, which explains its fundamental weakness across both protocols.

Figure 6 shows our method consistently achieves the highest user ratings across all three perceptual dimensions. Due to space limitations, the corresponding textual instructions are provided in the Appendix. Taken together, our quantitative metrics and subjective evaluations demonstrate that existing baselines struggle to produce natural, text-aligned choreography, often deviating from the real dance distribution. These findings underscore that current general-purpose models, constrained by a scarcity of dance-specific data, remain insufficient for fine-grained text-driven dance generation, thereby validating the effectiveness of our proposed approach.

Table 3: Quantitative ablation results evaluated on a 3,712-sample subset. We systematically ablate three core components: (1) the Choreographic Syntax, (2) the 3D motion representation, and (3) the components of the Choreographic Syntax. All variants share identical architectures and training configurations. For readability, AIST++ FID values are scaled by $10^{-1}$. ↓ or → denote whether a lower or closer-to-real value is optimal. Best results are bolded.

| Setting | HumanML3D Protocol | | | AIST++ Protocol | | | |
| --- | --- | --- | --- | --- | --- | --- | --- |
| | FID ↓ | MM Dist → | Diversity → | $FID_k$ ↓ | $FID_g$ ↓ | $Dist_k$ → | $Dist_g$ → |
| Real Data | - | 3.704 | 2.836 | - | - | 11.122 | 7.034 |
| w/o Choreographic Rules | 2.112 | 4.158 | 3.614 | 9.480 | 6.544 | 13.625 | 12.230 |
| w/o MHR (SMPL-X) | 2.799 | 3.487 | 2.795 | 12.055 | 10.446 | 7.036 | 10.177 |
| w/o Effort Dynamics | 1.030 | 4.002 | 3.208 | 0.667 | 2.126 | 9.741 | 8.502 |
| w/o Representation Refinement | 2.713 | 4.168 | 2.559 | 1.140 | 1.443 | 9.054 | 4.822 |
| **Full Model (Ours)** | **0.700** | **3.876** | **2.802** | **0.602** | **0.747** | **9.998** | **6.038** |

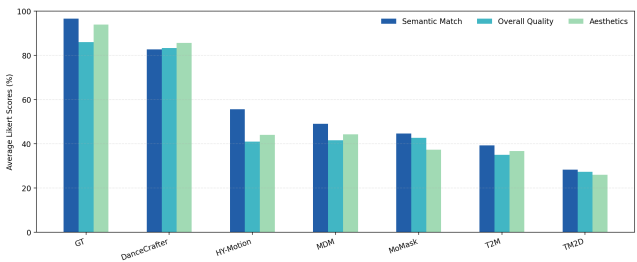

Figure 6: User study results for the main experiments.

## 5.3 Ablation Studies

To validate our key design choices, we conduct ablation studies on a hierarchically sampled subset of 3,712 sequences (3,500 for training, 212 for testing). To ensure fair comparisons, all model variants share identical architectures, hyperparameters, and training configurations. We systematically ablate four core components: (1) the Choreographic Syntax, (2) the 3D motion representation, (3) specific components within the syntax, and (4) the representation refinement. Quantitative results are provided in Table 3.

**The Choreographic Syntax.** To evaluate our annotation strategy, we introduce the "w/o Choreographic Rules" variant, which replaces our specialized syntax with basic text descriptions generated via Qwen3-VL-30B-A3B-Instruct [1]. These baseline prompts provide only coarse movement summaries, lacking formal choreographic structure and domain-specific vocabulary. As reported in Table 3, discarding these rules causes severe performance degradation. Under the HumanML3D protocol, FID worsens from 0.700 to 2.112, and MM Dist increases from 3.876 to 4.158. The impact is even more pronounced under AIST++: $FID_k$ surges from 0.602 to 9.480, and $FID_g$ jumps from 0.747 to 6.544. This confirms that simplistic descriptions fail to capture the spatial, temporal, and dynamic intricacies of professional dance, whereas our fine-grained annotations supply crucial supervisory signals.

**The Motion Representation.** To verify that MHR's decoupled skeletal and mesh representation is better suited for dance motion representation than the coupled SMPL-X formulation, we construct the "w/o MHR (SMPL-X)" variant by replacing MHR with SMPL-X parameters while keeping the text annotations unchanged. As Table 3 shows, discarding MHR causes consistent degradation. Under HumanML3D, FID increases from 0.700 to 2.799. Under AIST++, $FID_k$ spikes from 0.602 to 12.055, and $FID_g$ rises from 0.747 to 10.446. These results emphasize MHR's suitability for dance generation: its decoupled design affords a more continuous, coherent learning space, circumventing the parameter entanglement in SMPL-X that severely disrupts complex motion synthesis.

**The Effort Dynamics Component.** To isolate the impact of specific components within our Choreographic Syntax, we ablate the *effort dynamics* dimension. We filter out effort-related content in Choreographic Syntax, re-annotate the dataset via Gemini-3-pro-preview, and train the "w/o Effort Dynamics" variant. Removing effort dynamics incurs moderate but consistent performance drops. Under HumanML3D, FID increases from 0.700 to 1.030, and Diversity drifts from 2.802 to 3.208 (further from the 2.836 ground truth). Under AIST++, $FID_g$ rises from 0.747 to 2.126, and $FID_k$ slightly increases from 0.602 to 0.667. These findings underscore that effort dynamics are vital for capturing the geometric expressiveness and qualitative nuances of professional choreography.

**Representation Refinement.** We further ablate our continuous manifold representations and hybrid regularizations ("w/o Representation Refinement"). Table 3 shows removing them degrades quality and diversity. Detailed analysis is deferred to the Appendix.

## 6 Conclusion

To address the current scarcity of research and the absence of high-quality datasets in fine-grained text-to-dance generation, this paper proposes a comprehensive, full-stack solution encompassing theoretical foundations, data construction, and model design. **At the theoretical level**, we pioneer a cross-disciplinary integration of dance theory and anatomy to introduce a novel *Choreographic Syntax* alongside an annotation system, fundamentally resolving the difficulty of accurately and structurally describing dance motions. **At the data level**, we leverage this syntax to integrate professional academy archives with high-quality motion capture data. This culminates in the construction of **DanceFlow**, which, to the best of our knowledge, is the most fine-grained text dance dataset to date, establishing a robust foundation for future research. **At the model level**, we develop a generation model tailored to the dynamic characteristics of dance. By adopting the decoupled MHR, we achieve high-fidelity modeling of complex body parts. Finally, by cascading a video generation expert, our framework synthesizes highly expressive and photorealistic dance videos.

## A Dataset Details

### A.1 Motion Capture Workflow

To construct the high-fidelity portion of the **DanceFlow** dataset, we recorded professional dancers in a specialized motion capture laboratory using a Vicon optical motion capture system. The raw captured data consists of high-precision motion sequences in FBX format (containing joint coordinates, rotations, and skeletal hierarchies) and synchronized multi-view high-definition (HD) reference videos. Specifically, we recorded from three distinct perspectives—front, right-back, and left-back—at a resolution of 1080p and a frame rate of 60fps using the H.264 encoding format.

The raw FBX data is subsequently processed through a multi-stage pipeline: it is first converted into BVH skeletal motion files, then mapped to the SMPLX parametric human model, and finally transformed into MHR representation for downstream generative tasks. Figure 7 and Figure 8 provide visualizations of this specialized recording setup and the resulting 3D motion reconstructions, illustrating the rigorous foundation for our dataset.

### A.2 User Study Prompts and Recruitment

Due to space constraints in the main manuscript, we show the prompt for Figure 5 below:

> *Facing the 8 o'clock direction, the dancer balances on a single leg with the left leg supporting, the right leg lifted behind in a bent-knee attitude, foot fully pointed. The torso inclines forward toward 8 o'clock; the left arm curves naturally overhead, the right arm extends down and back, and the gaze is directed toward 8 o'clock. The weight then drops sharply as the right foot steps down firmly, the body rotating from 8 o'clock to face 1 o'clock while both knees bend deeply into a low squat. Simultaneously, the arms follow the turn, tracing a vertical circular pathway in front of the body, finishing in a low, grounded squat facing 1 o'clock with the torso tilted to the right; both arms are bent to frame the head, the left elbow lifted toward the upper left, the right elbow lowered toward the lower right, the backs of the hands close to the face in a cradling gesture. Immediately, the legs drive the body into a spiraling rise, the center lifting as the dancer pivots clockwise toward 8 o'clock on the left foot. The arms rotate with the body, opening from the chest and extending outward to the sides. The movement flows into a deep forward lunge facing 8 o'clock: the left leg bent and bearing weight in front, the right leg extended straight behind, the center slightly projected forward. The left arm reaches horizontally toward 8 o'clock with the wrist upright and palm pushing outward, the right arm extends back toward 4 o'clock with the palm down; the head turns toward 8 o'clock, the gaze following the left hand, settling into a "tailwind flag" pose.*

To assess the perceptual quality of the generated motions, we recruited 20 graduate students as evaluators. This group includes 5 professional graduate students from dance studies and 15 non-dance-related graduate students from various academic backgrounds. They independently rated each sequence on a 5-point Likert scale (1=worst, 5=best) based on three criteria: Text-Motion Alignment, Motion Quality, and Aesthetic Appeal.

### A.3 Expert Evaluation of Annotations and Annotation Interface

To guarantee the quality and accuracy of the 6.34 million text annotations in the **DanceFlow** dataset, we established a rigorous quality control mechanism. As shown in Fig. 9, we developed a dedicated annotation system in which expert evaluators can synchronously inspect each 3D dance motion sequence together with its corresponding machine-generated choreographic description across the four Syntax dimensions. The system supports both annotation revision and quality scoring, enabling experts to refine descriptions based on spatial precision, anatomical correctness, and dynamic effort while assigning an overall quality score.

To guarantee annotation fidelity at scale, we adopt a statistical quality control framework [17]. The 20K segments are divided into 100 batches, with $n$=30 samples per batch randomly drawn for review. Experts rate each sampled annotation on a 5-point scale, and scores below 3 are deemed unacceptable. Batches falling below a 95% acceptance rate are re-annotated until compliant. Fig. 13 presents the score distribution over 100 randomly sampled results from the final scored set.

### A.4 Dance Category Examples

To explicitly demonstrate the rich diversity and high-quality annotations of our **DanceFlow** dataset, we provide representative samples of various dance genres. Figure 10 and Figure 11 showcase excerpts and descriptions from Ballet, Breaking, Contemporary, Spanish, Dunhuang, Han-Tang, Shenyun, and Yangge.

## B Animation

Beyond 3D motion generation, **DanceCrafter** can further synthesize high-fidelity and expressive dance videos with the assistance of Wan-Animate [6]. This capability is especially important for choreography, as some dance movements are closely tied to costume design and character presentation, and therefore benefit from direct video-level visualization.

Our animation workflow is illustrated in Fig. 12. We first specify the desired movement using our Choreographic Syntax and generate the corresponding 3D dance motion with **DanceCrafter**. We then extract the skeletal motion sequence from the generated 3D motion and feed it, together with a reference image, into Wan-Animate. The video generation model animates the subject in the reference image according to the target dance motion, producing a photorealistic dance video that preserves both the intended choreography and the visual appearance of the reference character.

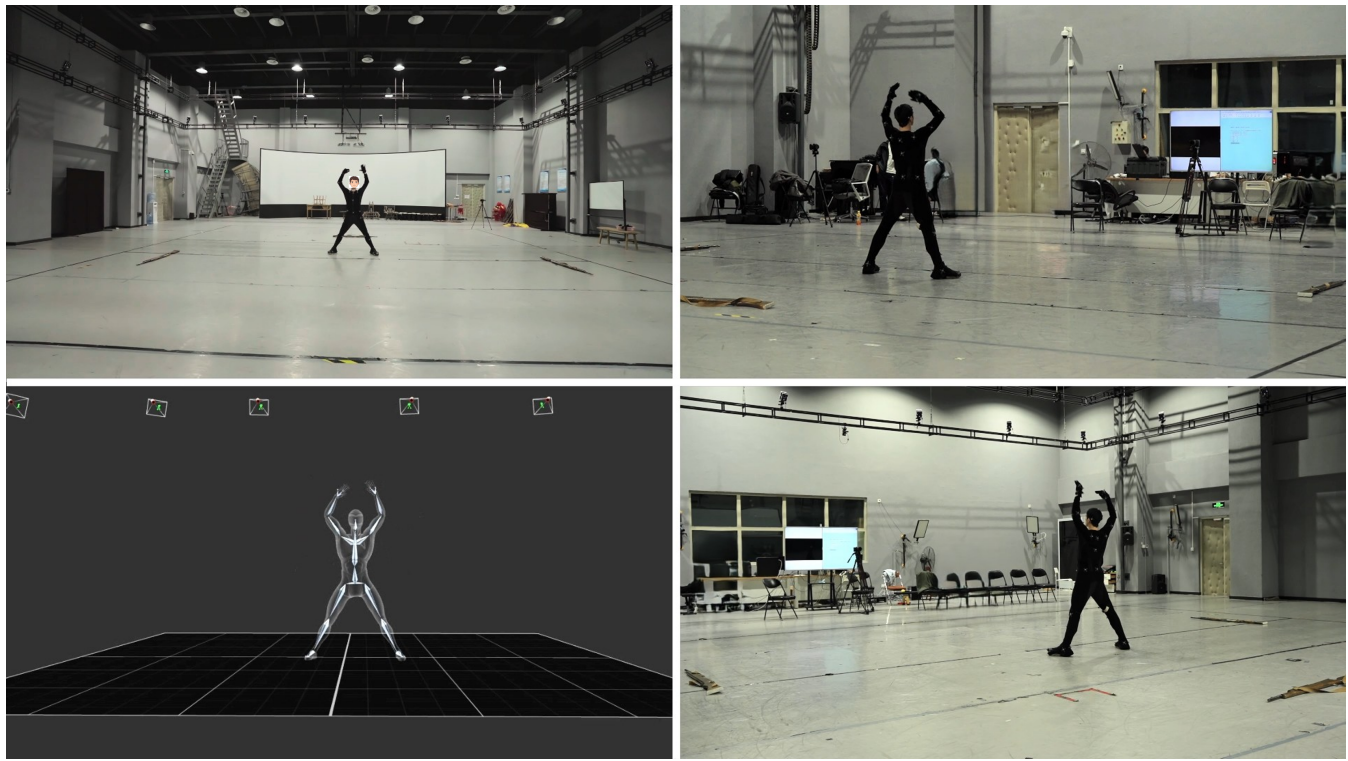

Figure 7: Overview of our professional motion capture recording for the DanceFlow dataset.

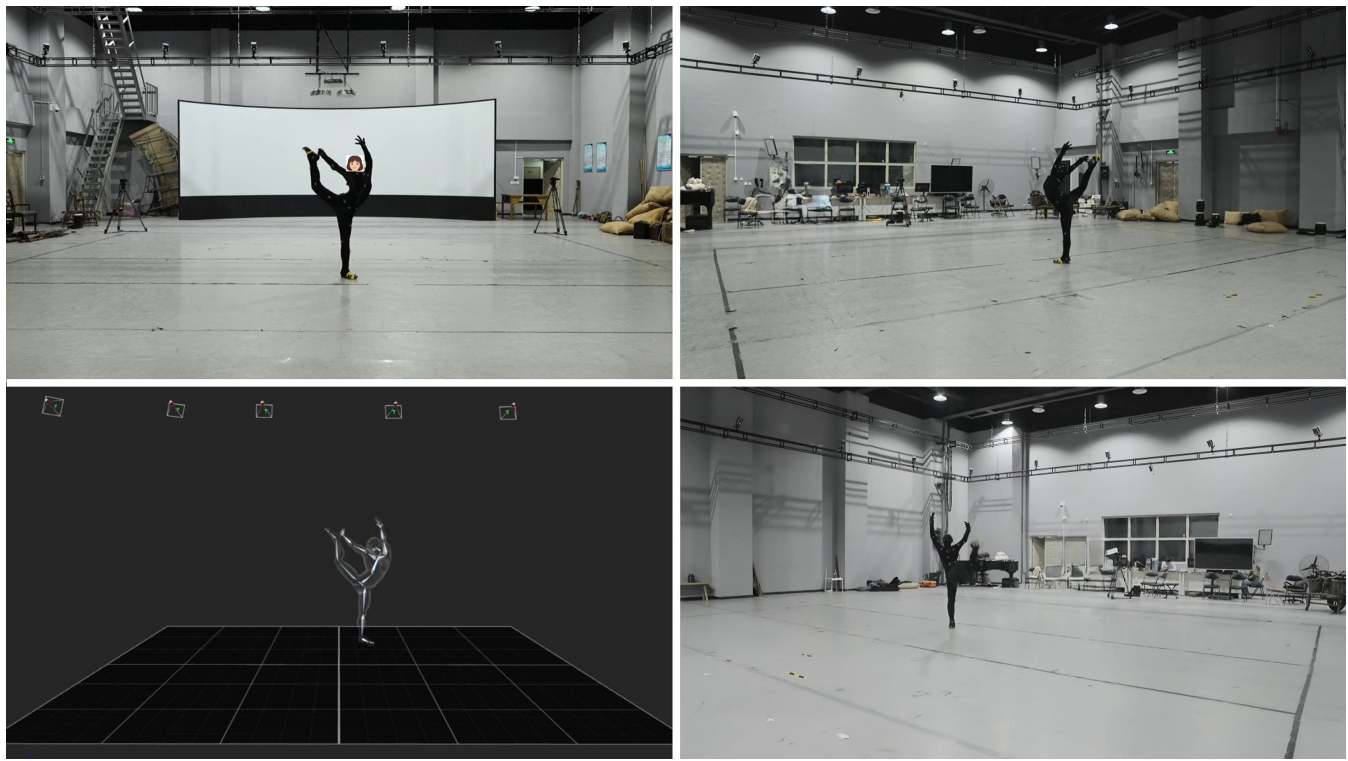

Figure 8: Detailed visualization of the motion capture data processing and 3D reconstruction results.

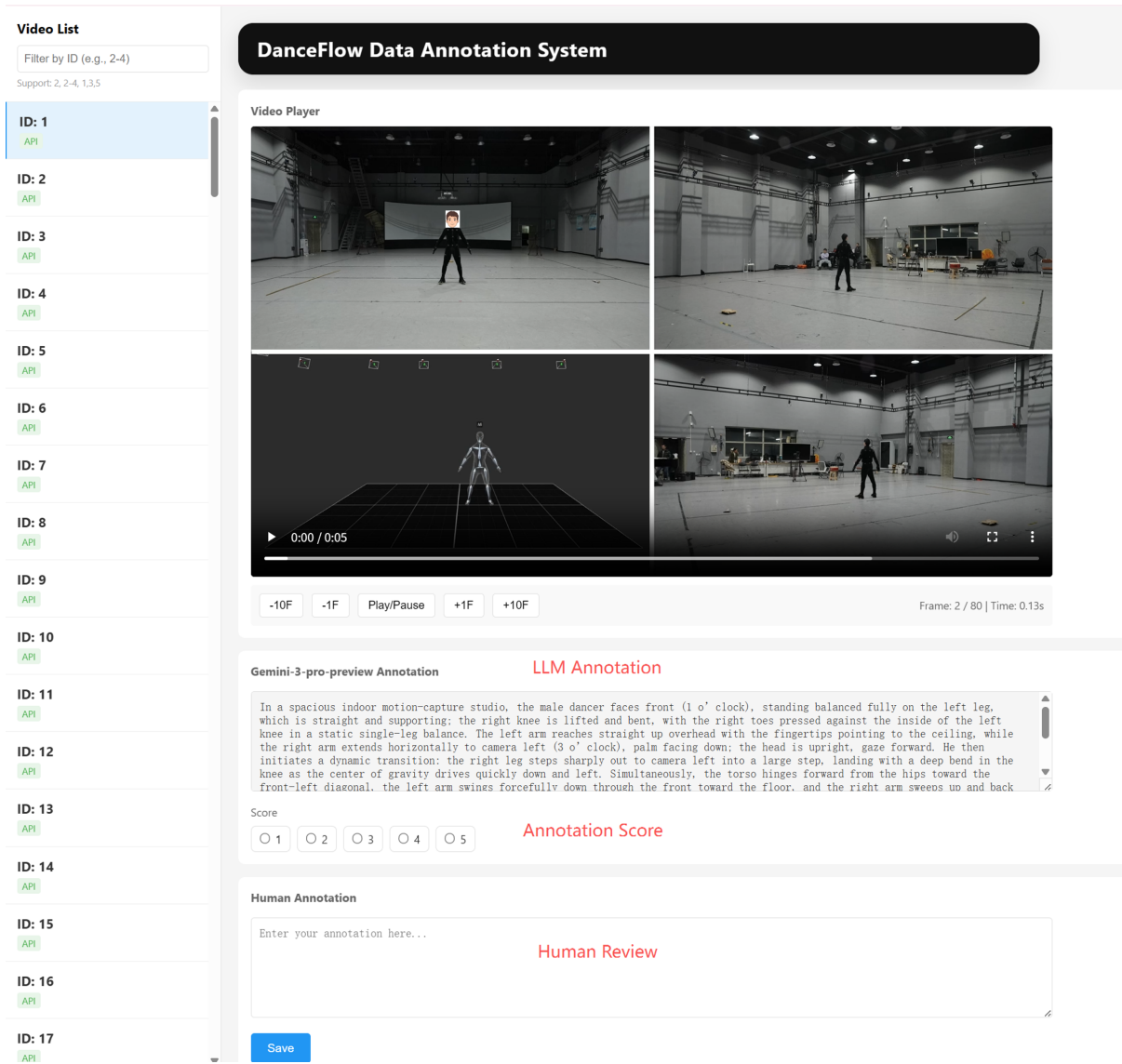

Figure 9: Overview of our specialized annotation interface used by domain experts to audit and refine the DanceFlow dataset.

## C Implementation Details

### C.1 Model Architecture and Training Recipe

Our generating backbone is a Diffusion Transformer (DiT), heavily optimized for the continuous 260-dimensional MHR manifold. For the main experiments, we employ a 12-layer Transformer architecture with a hidden dimension of 1024 and an FFN dimension of 4096. To encode the dense choreographic text, we utilize the pretrained UMT5-XXL text encoder, keeping its weights entirely frozen. Given the high temporal volatility of dance sequences, we apply Rotary Position Embeddings (RoPE) coupled with QK-Norm to stabilize the relative positional tracking across attention heads safely.

During the flow-matching training phase for the main experiments, we adopt AdamW with a learning rate of $1 \times 10^{-4}$, batch size 16, dropout 0.05, conditioning drop probability 10%, and EMA decay 0.9999. Training is conducted on 8 A100 GPUs for 250K steps. The loss weights are set to $\lambda_{\text{rot}} = 1.0$, $\lambda_{\text{body}} = 1.5$, $\lambda_{\text{hand}} = 0.5$, $\lambda_{x_0} = 2.0$, $\lambda_v = 0.5$, and $\lambda_a = 1.5$. At inference time, we employ a 50-step Euler integrator with Classifier-Free Guidance (CFG) using a guidance scale of $w = 1.0$.

For the ablation studies, we adopt a lighter DiT configuration with hidden dimension 512 and FFN dimension 2048. We train with AdamW using a learning rate of $2 \times 10^{-4}$, batch size 8, dropout 0.1, conditioning drop probability 10%, and EMA decay 0.9999 for 30K steps. Unless otherwise specified, we keep the remaining loss and inference hyperparameters aligned with the main setting, namely

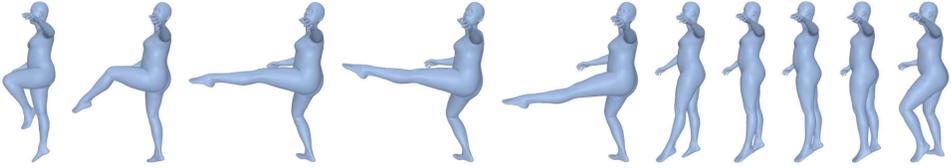
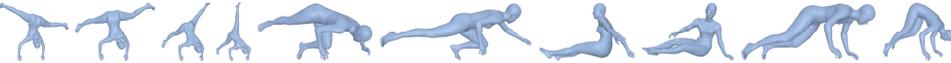
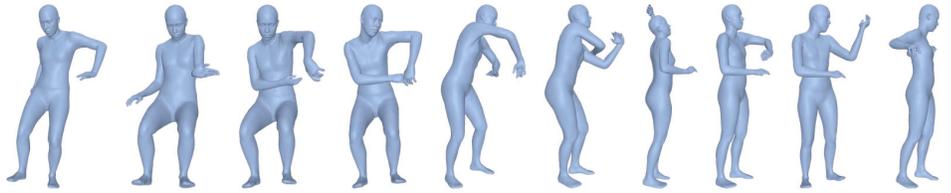
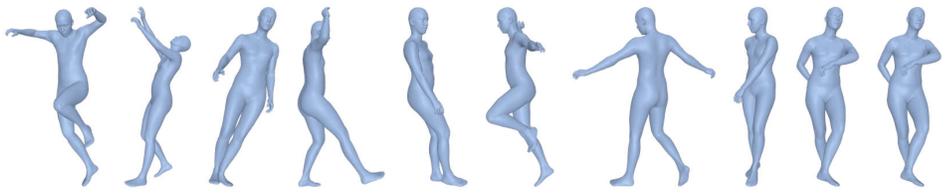

**Figure 10: Representative 3D motion excerpts and fine-grained descriptions of Ballet, Breaking, Contemporary, and Spanish dance from the DanceFlow dataset.**

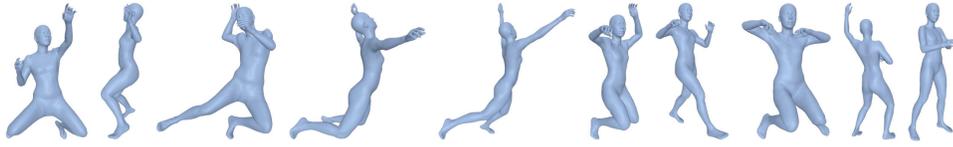

**Modern:** The dancer begins in a low, grounded position on the floor, facing the right side of the video (7 o'clock), with the legs crossed or kneeling and concealed by the skirt. Engaging the core and pushing through the lower body, the dancer initiates a counterclockwise turn, spiraling upward as the body rises from a grounded level to an upright stance, completing approximately a 360-degree rotation in one continuous, rapid motion. The movement resolves abruptly into stillness, finishing in a standing position facing the right front of the video (8 o'clock). The weight is evenly placed on both feet, the torso subtly arching backward to lengthen the back line. The right hand holds a fan, the arm fully extended toward the upper right (2 o'clock high), while the left arm lowers naturally toward the lower left (6 o'clock low) with the palm pressing downward, creating a strong diagonal line through the arms. The head tilts upward, the gaze following the raised fan, forming an erect, expansive, and composed final pose.

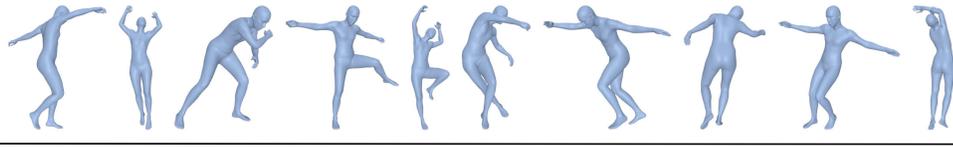

**Dun Huang:** The dancer lands from an aerial jump, dropping the center of gravity sharply into a low side lunge: the right leg (screen right) bends to bear weight while the left leg (screen left) extends long to the left, skimming the floor; the torso inclines forward-left, and both arms hang naturally with the momentum to maintain balance. The center then rebounds quickly upward and shifts onto the right leg, which straightens to support as the left knee lifts, the foot pointed and lightly turned inward, placed near the inside of the right knee in a one-legged balance; the upper body tilts and twists to the left, with both elbows bent to form a rounded, embracing shape above the left side of the head, the gaze dropping diagonally down-left. Immediately, the left foot steps across to the right, carrying the weight rapidly to the right as the body reaches forward-right; the right arm extends forcefully to the right at shoulder height while the left arm stretches back, creating an expansive running dynamic. Riding this momentum, the dancer pivots on the right foot, executing one full clockwise turn to face front (1 o'clock), with both arms sweeping from in front of the chest outward to a horizontal open position. The movement stops abruptly: the feet stand apart, arms extended straight to each side in a strong horizontal line with palms down, and the head snaps sharply to the left, eyes looking level toward stage left (3 o'clock), forming a tense, charged still pose.

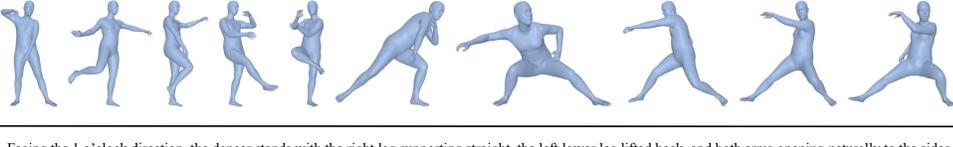

**Shen Yun:** Facing the 1 o'clock direction, the dancer stands with the right leg supporting straight, the left lower leg lifted back, and both arms opening naturally to the sides. The left foot then steps to the 3 o'clock direction, shifting weight left, as the right foot crosses forward to 2 o'clock in front of the left; both knees soften and the torso inclines slightly forward. The arms cross and sweep in front of the chest with the steps. Next, pivoting on the right foot, the left knee lifts in to a passé position with the foot pointed near the right knee, and the body turns one full rotation counterclockwise. During the turn, the right arm bends to guard in front of the chest while the left arm reaches upward. Finishing the turn, the left foot quickly strides out to 3 o'clock, the left knee deeply bent and the right leg extended toward 7 o'clock, dropping sharply into a side lunge. Simultaneously, the torso stretches toward 3 o'clock, both arms snap open strongly to a horizontal line—left arm pointing to 3 o'clock, right arm to 7 o'clock—with palms open, the head whipping to the left to look along the left arm, ending in a held pose with clear accents and power.

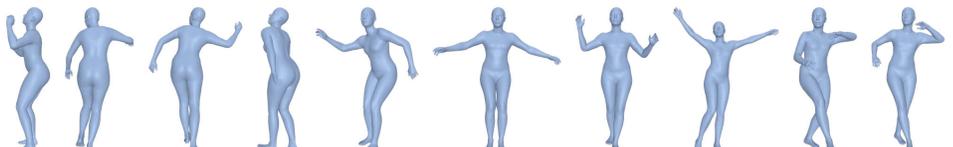

**Chinese Folk:** The dancer begins in a still pose with the back to the audience, facing the 5 o'clock direction, weight on the right leg, left toe lightly touching the floor behind, head slightly inclined to the right, arms relaxed and hanging by the sides. Using the right foot as a pivot, the dancer quickly turns clockwise 180 degrees to face 1 o'clock; propelled by the centrifugal force, both arms fling open horizontally to the sides as the long scarves fly outward. The dancer briefly holds a standing position facing 1 o'clock, feet set slightly wider than shoulder width, knees softly bent, arms extended sideward at shoulder height, the blue handkerchief in the right hand pointing right and the pink handkerchief in the left hand pointing left. The movement then flows into a continuous left–right sway: the weight shifts left, the left knee bends and the right leg lengthens, the torso tilts left as the right arm arcs overhead to cover toward the left. The weight then smoothly transfers to the right, the right knee bending to support while the left leg extends to the side with the toe touching the floor; the torso inclines to the right, stretching the left side of the waist. Simultaneously, the left arm sweeps across the chest and arcs upward to the left high diagonal, while the right arm circles low to the right, palm pressing to the right waist; the head turns to the right, gaze directed toward the 8 o'clock direction, finishing in a right-weighted, side-tilted pose facing 1 o'clock.

**Figure 11: Additional dance genre examples including Modern, Dunhuang, Shenyun, and Yangge, showcasing the diversity of our dataset.**

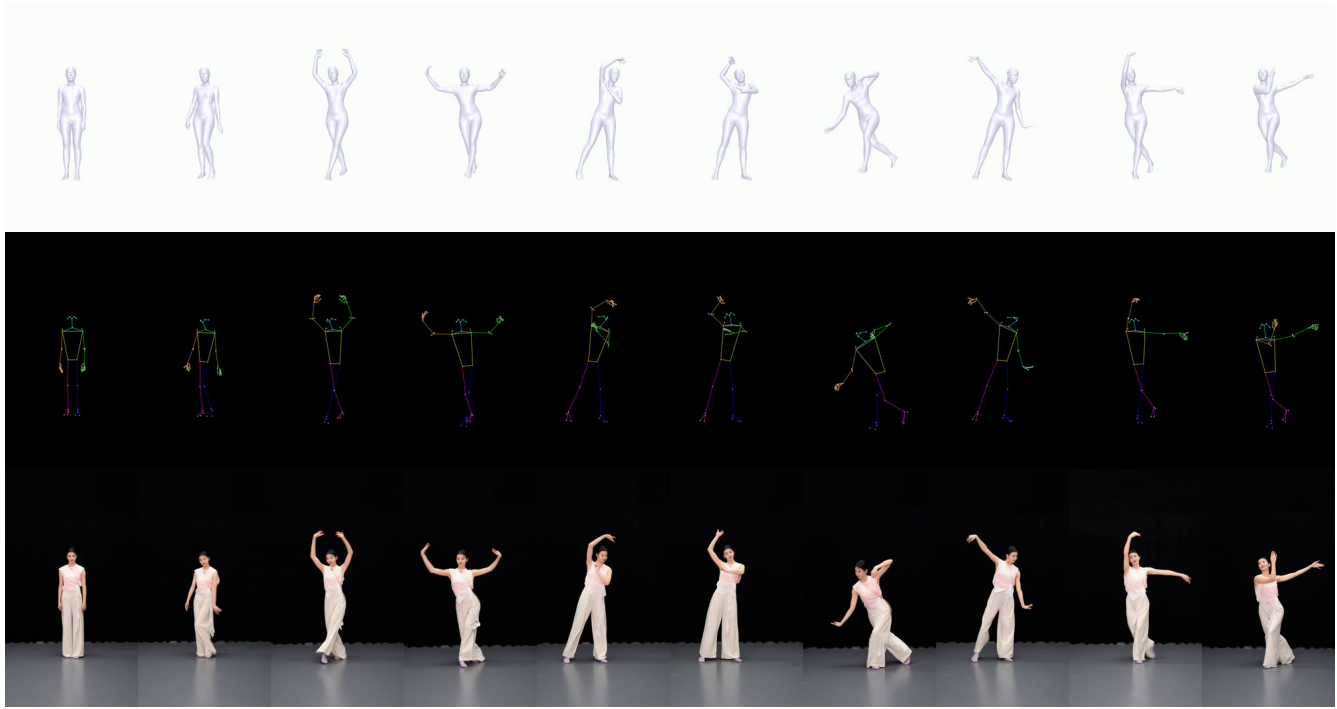

Figure 12: Animation workflow and generation examples. Given a choreographic description specified with our Choreographic Syntax, DanceCrafter first generates a 3D dance motion. The corresponding skeletal motion, together with a reference image, is then fed into Wan-Animate to synthesize a photorealistic and expressive dance video.

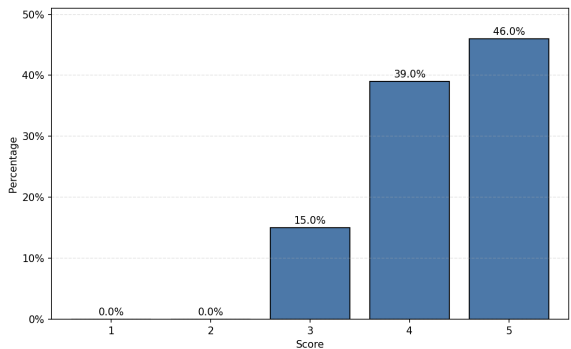

Figure 13: Distribution of expert quality scores over 100 randomly sampled annotated results.

$\lambda_{\text{rot}} = 1.0$, $\lambda_{\text{body}} = 1.5$, $\lambda_{\text{hand}} = 0.5$, $\lambda_{x_0} = 2.0$, $\lambda_v = 0.5$, $\lambda_a = 1.5$, and guidance scale $w = 1.0$.

## C.2 Representation Refinement Ablation

Beyond the main ablations, we detail the experimental setup for evaluating our continuous manifold data representation and hybrid normalization strategy. In the "w/o Representation Refinement" variant (Table 6), we directly regress the native 136-dimensional MHR pose parameters without applying any $\mathbb{R}^6$ or sine-cosine manifold conversions. Consequently, since the representations no longer reside on specialized geometric manifolds, we also abandon the hybrid regularization mechanism. Instead, we apply a standard $z$-score normalization—subtracting the mean and dividing by the standard deviation—uniformly across all 136 parameter dimensions.

As demonstrated in Table 6, forcing the network to directly learn these discontinuous parameter spaces notably worsens all generation metrics. Beyond the quantitative decline, practical qualitative observations reveal that the generated dance motions suffer from severe instability. Specifically, the network frequently struggles with topological boundary wrap-arounds, resulting in abnormal visual jittering, unphysical body twisting, and overall structural collapses. These findings explicitly underscore the synergistic necessity of continuous motion representations alongside dedicated scale-preserving regularizations for high-fidelity dance generation.

## D Choreographic Syntax Framework

We present the complete and detailed Choreographic Syntax framework in the final part of the Appendix. With this syntax, dance movements can be described in a structured and standardized manner by following a unified annotation procedure.

# Choreographic Syntax

The minimum descriptive units of dance are the **"Static Pose"** and the **"Dynamic Connection"**.

- **Static Pose**: Refers to a state where there is no significant movement in any part of the dancer's body. For static poses, it is only necessary to describe the external form of the posture. *(Example: The arm is bent at approximately 90 degrees, the elbow is slightly raised, and the forearm naturally forms a slanted line.)*
- **Dynamic Connection**: Refers to the dancer's process of movement. When describing a dynamic connection, the following four aspects must be addressed:
    1. Body parts involved in the movement.
    2. Changes in the center of gravity.
    3. Spatial trajectory of the movement.
    4. Movement dynamics (force and speed).

Below is an introduction to these four components and their descriptive standards.

> **IMPORTANT NOTE ON PERSPECTIVE**: In the standards below, any reference to "Left" or "Right" is strictly based on the **Camera Perspective** (i.e., the viewer's/video's left and right), NOT the dancer's own anatomical left or right.

## 1. Body Parts

### 1.1 Head

Human movement often originates from the head. The **eyes** are the core, capable of guiding the visual focus of the overall movement.

- **Eyes**:
    - *Directions*: Up, Down, Left, Right, Diagonal-Up, Diagonal-Down.
    - *Description*: Used to guide the movement's line of sight or act as the focal point of the head posture.

### 1.2 Upper Limbs

The upper limbs encompass the shoulders, upper arms, elbows, forearms, and hands.

- **Shoulder (Left/Right)**: The foundation of upper limb movement, connecting to the upper arms.
- **Upper Arm (Left/Right)**: Connects the shoulder to the elbow.
- **Elbow (Left/Right)**: The critical joint for adjusting arm length and angle, extending downwards to the forearm.
- **Forearm (Left/Right)**: Connects the elbow to the wrist, frequently involved in rotational movements.
- **Hand (Left Hand / Right Hand)**: The extremity of the upper limb.
    - *Sub-parts*: Heel of the palm, center of the palm, fingers.
    - *Description*: Used to execute subtle, distal movement expressions.

### 1.3 Trunk

The trunk is the power center of the body.

- **Back**: The primary exertion side that supports body posture. *(Note: Left/Right here generally refers to the exertion side of the back muscles).*
- **Waist**: The central hub for core power transmission, body control, and twisting movements.
    - *Sub-parts*: Front waist (abdomen/core), Side waist (Left/Right), Lower back.
- **Abdomen**: Achieves precise control over lateral movement and balance (e.g., via the external oblique muscles).

### 1.4 Lower Limbs

The lower limbs bear the responsibility of support and displacement.

- **Hips/Crotch**: The root of all lower limb movements and the key to shifting the center of gravity. *(Directions: Front, Back, Left, Right, Up, Down)*
- **Thigh**: One of the primary sources of power. *(Directions: Front, Back, Left, Right, Up, Down)*
- **Knee**: The crucial joint for weight-bearing and shock absorption. *(Directions: Front, Back, Left, Right, Up, Down)*
- **Calf**: Connects the knee to the ankle. *(Directions: Front, Back, Left, Right, Up, Down)*
- **Ankle**: The joint that determines balance and flexibility. *(Directions: Lift/Relevé, Drop/Plié)*
- **Foot (Left Foot / Right Foot)**: The critical distal part for cushioning and exerting force upon landing.
    - *Sub-parts*: Ball of the foot, arch, heel, toes.

## 2. Center of Gravity (COG)

### 2.1 State of Gravity

- **Single-Leg COG**: Full-foot weight or half-foot (relevé) weight.
- **Two-Leg COG**: One supporting leg acts as the primary pillar, while the working leg acts as an auxiliary.
- **Opposing COG**: Jumping/Airborne (suspended in the air).

### 2.2 Changes in Gravity

- **Maintain**: Keeping the center of gravity unchanged.
- **Shift/Push**: Transferring weight, usually involving a process of sinking the center of gravity.
- **Slow Transfer**: Moving the weight at a slower speed.
- **Instant Switch**: Transferring weight rapidly.

# 3. Spatial Trajectory

## 3.1 Classification of Body Movement Space

The human body moves within three primary planes:

| Space Classification | Anatomical Name | Spatial Description | Movement Logic & Examples |
|---|---|---|---|
| **Table Plane** | Transverse Plane | Divides the body into **Top** and **Bottom** halves. *Imagine a giant round table around your waist. Movements can only slide along the tabletop.* | **Keywords:** Rotate, Twist. Movements revolve around the Vertical Axis (horizontal arcs/circles). *Examples:* Ballet pirouettes, horizontal torso translation, waist twists. |
| **Wheel Plane** | Sagittal Plane | Divides the body into **Left** and **Right** halves. *Imagine you are a giant wheel rolling forward or backward in a narrow hallway.* | **Keywords:** Front, Back, Flexion, Extension. Movements revolve around the Coronal Axis (forward/backward lines/arcs). *Examples:* Forward rolls, backbends, forward leg stretches. |
| **Door Plane** | Frontal (Coronal) Plane | Divides the body into **Front** and **Back** halves. *Imagine standing in a door frame. Movements must stay flat against the wall behind you like windshield wipers.* | **Keywords:** Left, Right, Side-bend, Open/Close. Movements revolve around the Sagittal Axis (left/right lines/arcs). *Examples:* Cartwheels, side stretches. |

## 3.2 Spatial Directions

- **Basic Dimensions**: Up - Down, Left - Right, Front - Back.
- **Complex Dimensions**:
    - *Table Plane Diagonals*: Front-Left, Front-Right, Back-Left, Back-Right.
    - *Door Plane Diagonals*: Top-Left, Top-Right, Bottom-Left, Bottom-Right.
    - *Wheel Plane Diagonals*: Front-Up, Back-Up, Front-Down, Back-Down.

## 3.3 Objective Space (The Stage)

Stage orientation uses an "8-o'clock System," with the dancer as the center point facing the camera. If a direction is ambiguous, describe it using the nearest point.

| Point | Spatial Direction | Description |
|---|---|---|
| **1 o'clock** | Direct Front | Facing the audience (the exact position of the camera). |
| **2 o'clock** | Front Left | The 45-degree angle between Point 1 and Point 3. |

| Point | Spatial Direction | Description |
|---|---|---|
| **3 o'clock** | Direct Left | The direct left side of the video frame. |
| **4 o'clock** | Back Left | The 45-degree angle between Point 3 and Point 5. |
| **5 o'clock** | Direct Back | Opposite the camera; the dancer's back is to the audience. |
| **6 o'clock** | Back Right | The 45-degree angle between Point 5 and Point 7. |
| **7 o'clock** | Direct Right | The direct right side of the video frame. |
| **8 o'clock** | Front Right | The 45-degree angle between Point 7 and Point 1. |

## 4. Movement Dynamics

### 4.1 Movement Speed

**Objective Duration (Length of Time):**

- **Instantaneous**: Extremely short completion time, process barely visible (<0.5 seconds).
- **Brief**: Short duration, but the process is perceptible.
- **Moderate**: A speed that aligns with natural, everyday rhythms.
- **Prolonged**: Deliberately lengthened movement duration, exceeding natural breathing rhythm.
- **Frozen**: Speed is zero, maintaining a static state for a certain duration.

**Speed Change Process:**

- **Constant**: Unchanging speed, displaying a strong sense of mechanics or control.
- **Accelerating**: Gradually increasing speed, feeling like a release after gathering energy.
- **Decelerating**: Gradually decreasing speed, usually accompanied by a fading of force.
- **Sudden/Explosive**: Reaching peak speed in a fraction of a second without warning.
- **Abrupt Stop/Brake**: Instant stillness during high-speed movement, emphasizing immense control.
- **Cushioning**: Transitioning from fast to slow, absorbing force (e.g., upon landing).
- **Hovering/Suspended**: Seemingly still at the peak or a certain point, but with internal extension.
- **Trembling**: Extremely high-frequency, small-amplitude, rapid back-and-forth movements.

### 4.2 Movement Intensity (Force)

| Level | Descriptive Terms | Meaning & Visual/Physical Characteristics |
|---|---|---|
| **Extreme** | Full power / Limit | Power bordering on physical limits. Muscles extremely visibly tense, possibly accompanied by shaking; carries a sense of destruction. |
| **Strong** | Heavy / Forceful | Fully utilizing body weight and gravity. Visually solid, like a falling rock; difficult to push. |

| Level | Descriptive Terms | Meaning & Visual/Physical Characteristics |
|---|---|---|
| **Med-Strong** | Full / Solid | Muscles fully activated, strong support. Clear body lines without sagging; energy fills the entire silhouette. |
| **Medium** | Moderate / Normal | Baseline force for daily life or standard movements. Neither strenuous nor slack; often ignored because it is too "normal." |
| **Med-Weak** | Light / Nimble | Just enough force to counteract gravity, with a slight upward lift. Visually like a floating balloon. |
| **Weak** | Slight / Minimal | Retaining minimal muscle control. Feels like merely resting on an object with only skin contact. |
| **None** | Relaxed / Loose | Completely abandoning active force (drooping via gravity). Visually like wet clothes on a hanger; entirely passive. |

## 4.3 Movement Fluency

- **Non-fluent**: Stuttering / Sluggish / Viscous
- **Normal**: Continuous / Unbroken
- **Fluent**: Smooth / Linear

## 4.4 Composite Descriptions

*Formula: Movement Speed + Exertion Method*

- **Sluggish / Viscous**: Moving as if through glue; slow with resistance. Not just slow, but heavy.
- **Ethereal / Floating**: Moderate to fast speed, but with very weak gravitational pull. Like a falling feather.
- **Agile / Crisp**: Fast reaction, short and clean movement paths, no dragging.
- **Passing / Gliding**: Emphasizes the path of movement; speed is usually even and smooth.
- **Striking / Impacting**: Extremely fast with a strong sense of sudden stopping/impact at the endpoint.
- **Continuous / Extending**: Slower speed, but with a continuous sense of propulsion; no breakpoints.

# 5. Description Format and Examples

**Structural Order**: Always describe the **Environment and Character** first, followed by the **Action Description**.

## 5.1 Environment and Character

1. Describe the environment setting.
2. Describe the character's clothing and appearance.

## 5.2 Action Description

Determine if it is a Static Pose or a Dynamic Connection.

- **Static Pose**: Briefly describe the external form of the posture.
- **Dynamic Connection**: First check for gravity changes, then identify the body parts, trace the spatial path, and finally supplement with movement dynamics.
- *Note: Static poses and dynamic connections do not appear in a fixed order; they may alternate, or a sequence may contain only dynamic movements.*

## Crucial Rules for Description

1. **Do not over-detail**: Spatial paths should not be described with excessive micro-details. Ignore negligible micro-movements.
2. **No subjectivity**: Subjective aesthetic adjectives or anthropomorphic metaphors are strictly forbidden.
3. **Seamless integration**: Do not artificially label "Static Pose" or "Dynamic Connection" in your output text. If it is static, describe the pose; if it is dynamic, detail the action. Blend them seamlessly into cohesive paragraphs.
4. **Camera Perspective**: When describing Left and Right, always use the **Camera Perspective** (the video's left/right), NOT the dancer's own left/right. Do not explicitly write "the video's left"; simply state "Left" or "Right."

## 5.3 Example Description

> A barefoot male dancer wearing a dark green slanted-placket long shirt and black wide-leg pants stands in the center of the stage against a black background and a grey striped floor.
>
> The dancer stands facing point 6, with his back to the audience, his feet slightly wider than shoulder-width apart, his head slightly bowed, and his arms hanging naturally. Using his left foot as an axis, his body turns clockwise from point 6 to point 1. His right leg brushes the floor, stepping back and to the side. Both knees bend, and his center of gravity presses down. Simultaneously, his arms cross in front of his abdomen; his right arm extends diagonally upward to the right, and his left arm extends diagonally downward to the left. His elbows are straight, palms open, and his head turns towards point 1, looking at his right hand. He freezes in a low-gravity deep lunge facing point 1, with his right arm diagonally up and his left arm diagonally down. Subsequently, his legs straighten, lifting his body, and his center of gravity shifts to the center of both feet. His arms gather from the sides toward his chest, wrists touching and turning clockwise, causing his five fingers to spread open. Immediately after, his arms sweep horizontally from his chest out to the sides, palms turning downward. He pauses, standing upright facing point 1, arms raised horizontally to the sides, head tilted slightly up. His right toe brushes the floor, drawing close to his left foot, then his right knee lifts. His right instep points straight, resting against the inside of his straight left knee. Simultaneously, his right arm traces an upward arc from the side to the 5th position above his head, palm facing up; his left arm cuts downward from the side to the front of his abdomen, fingertips pointing to point 3. He holds a one-legged upright pose with the working leg lifted, facing point 1, right hand held high supporting an invisible weight, left hand sweeping horizontally across the abdomen. Next, his right foot steps toward point 2. His left knee bends deeply as his center of gravity shifts forward and presses down, his torso leaning forward toward point 2. His right arm follows, cutting downward from the high position past the front to the floor, with his right palm hovering just above the ground. His left arm pulls diagonally upward and backward toward point 8, forming a low-gravity deep

> lunge facing point 2, right hand reaching for the ground, left hand pulling back. Keeping a low center of gravity, his body turns counter-clockwise from point 2 to point 8. His left knee turns inward and kneels on the floor, supported by the ball of his right foot. As his body turns, his right hand draws a horizontal arc forward from the ground, his left hand retracts, and finally, his arms cross over his chest, hugging his shoulders. He remains still in a kneeling posture facing point 8, arms hugging his shoulders, head bowed down. The dancer then stands up, his body turning clockwise to face point 5, back to the audience, feet together, arms hanging down.